\documentclass{sig-alternate}
\usepackage{mathptmx} 

\usepackage{paralist}
\usepackage{fancyhdr}
\usepackage[normalem]{ulem}
\usepackage[hyphens]{url}
\usepackage[sort,nocompress]{cite}
\usepackage[final]{microtype}
\usepackage[keeplastbox]{flushend}
\usepackage[bookmarks=true,breaklinks=true,letterpaper=true,colorlinks,linkcolor=black,citecolor=blue,urlcolor=black]{hyperref}
\usepackage{multirow}
\usepackage{subfigure}
\usepackage{algorithm}
\usepackage{algorithmic}
\usepackage{enumitem}
\usepackage{comment}
\usepackage{wrapfig}

\fancypagestyle{firstpage}{
  \fancyhf{}
  
  \fancyhead[C]{\vspace{15pt}\normalsize{}} 
  \fancyfoot[C]{\thepage}
}

\vspace{-0.5cm}
\title{AutoFL: Enabling Heterogeneity-Aware Energy Efficient Federated Learning
\vspace{-1.5cm}
} 

\author{
\;\; \normalsize{Young Geun Kim$^{\ast}$} \;\;\;\;\;\;\;\;\;\;\;\;\;\;\;\;\;\;\; \normalsize{Carole-Jean Wu$^{\dagger \delta}$} \\
\affaddr{\normalsize{Soongsil University$^{\ast}$}} \;\; \affaddr{\normalsize{Arizona State University$^{\dagger}$}} \;\; \affaddr{\normalsize{Facebook AI Research$^{\delta}$}}}

\begin{document}
\maketitle
\thispagestyle{firstpage}
\pagestyle{plain}


\begin{abstract}
Federated learning enables a cluster of decentralized mobile devices at the edge to collaboratively train a shared machine learning model, while keeping all the raw training samples on device. This decentralized training approach is demonstrated as a practical solution to mitigate the risk of privacy leakage. However, enabling efficient FL deployment at the edge is challenging because of non-IID training data distribution, wide system heterogeneity and stochastic-varying runtime effects in the field. 
This paper jointly optimizes time-to-convergence and energy efficiency of state-of-the-art FL use cases by taking into account the stochastic nature of edge execution. We propose AutoFL by tailor-designing a reinforcement learning algorithm that learns and determines which \textit{K} participant devices and per-device \textit{execution targets} for each FL model aggregation round in the presence of stochastic runtime variance, system and data heterogeneity. By considering the unique characteristics of FL edge deployment judiciously, AutoFL achieves 3.6 times faster model convergence time and 4.7 and 5.2 times higher energy efficiency for local clients and globally over the cluster of \textit{K} participants, respectively. 

\end{abstract}

\section{Introduction}
\label{sec:introduction}

The ever-increasing computational capacities and efficiencies of smartphones have enabled a large variety of machine learning (ML) use cases at the edge~\cite{CJWu2019}, such as image recognition~\cite{Google_1}, virtual assistant~\cite{Amazon,Apple}, language translation~\cite{Google_2}, automatic speech recognition~\cite{Google_3}, and recommendation~\cite{WJiang2020}. As the mobile ML system stack matures~\cite{Android,Nvidia,PyTorch2,Qualcomm,Samsung2,TChen2018,Tensorflow}, on-device inference becomes more efficient with innovations in algorithmic optimizations~\cite{KHe2017,ALavin2017,MMathieu2014,MSandler2018,MTan2019,BZoph2018,wu2018fbnet}, neural network architecture optimizations~\cite{AHoward2019,MSandler2018,ZSun2020,BWu2019}, and the availability of programmable accelerators~\cite{Apple2,Huawei,Huawei2,Google_4,Qualcomm,Samsung3,Samsung4}. While on-device inference is becoming more ubiquitous~\cite{ECai2017,AEEshratifar2020,MHan2019,YKang2017,VJReddi2020,VJReddi2021,NDLane2016,SWang2020_2,YGKim2020,CJWu2019,GZhong2019}, performing ML model training in the cloud is still the standard practice for most use cases~\cite{BAcun2021,AEEshratifar2020,KHazelwood2018,NPJouppi2017,PMattson2020,MNaumov2020,VJReddi2021}, due to the significant computation and memory resource requirements~\cite{AAAwan2017,ZGong2020,DLepikhin2020,SRajbhandari2021,YEWang2020,YEWang2020_2,CYin2021,BZheng2020}. 

Recently, Federated Learning (FL) enables smartphones to collaboratively train a shared ML model, while keeping all the raw data on device~\cite{KBonawitz2019,RCGeyer2017,SItahara2020,JKonecny2016,ALi2020,WYBLim2020,HBMcMahan2017,ZTao2018,CWang2021,YZhan2020,JZhang2020}. This decentralized training approach is a practical solution to mitigate the risk of privacy leakage in Deep Neural Network (DNN) model training, as only the model gradients, not individual data samples, are sent to update the shared model in the cloud~\cite{TSBrisimi2018,AHard2018,TLi2020}. The shared model is trained iteratively with the model gradients from a large collection of participating smartphones. While FL has shown great promise for privacy sensitive applications, such as sentiment learning, next word prediction, health monitoring, and item ranking~\cite{KBonawitz2019,AHard2018,ALi2020}, its deployment is still in a nascent stage. 

To enable efficient FL deployment at the edge, maximizing the computation-communication ratio by having less number of participant devices with higher per-device training iterations is a common practice for FL~\cite{HBMcMahan2017, VSmith2017, TLi2020}. In particular, FedAvg has been considered as the de facto FL algorithm~\cite{JKonecny2016,HBMcMahan2017}. At each aggregation round, FedAvg trains a model for \textit{E} epochs using Stochastic Gradient Descent (SGD) with minibatch size of \textit{B} on \textit{K} selected devices, where \textit{K} is a small subset of \textit{N} devices participating in the FL. The \textit{K} devices then upload the respective model gradients to the cloud where the gradients get averaged into the shared model. By allowing lower \textit{K}, FedAvg significantly reduces the amount of data transmission for each aggregation round. Various previous works have been also proposed to improve the accuracy of trained models~\cite{MDuan2021,ALi2020,TLi2020_2} or security robustness~\cite{RCGeyer2017,YLi2018,YLu2020} on top of the FedAvg algorithm.

\begin{figure}[t]
    \centering
    \includegraphics[width=\linewidth]{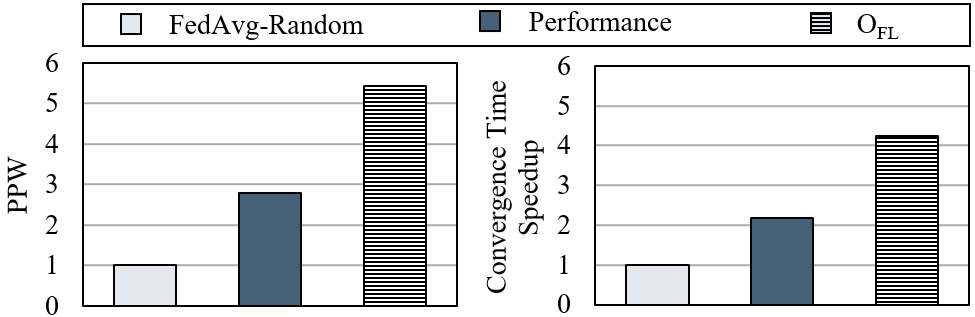}
    \vspace{-0.3cm}
    \caption{The performance-per-watt (PPW) energy efficiency of FL execution can be significantly improved by up to 5.4x with judicious selections of participant devices and the execution targets ({\tt Performance} and $O_{FL}$ --- Section~\ref{sec:methodology} for methodology detail).}
    \label{FL-motivation-oracle}
\end{figure}

While these advancements open up the possibility of efficient FL deployment, a fundamental challenge remains---deciding which \textit{K} devices to participate in each aggregation round for a given (\textit{B, E, K})\footnote{FL global parameters of (\textit{B, E, K}) are usually determined by the service providers considering the service-level accuracy requirements, and computation- and memory-capabilities of edge devices~\cite{KBonawitz2019,JKonecny2016}.}, and deciding which \textit{execution target} to perform model training on a participating device. State-of-the-art approaches randomly select \textit{K} participants from a total of \textit{N} devices~\cite{KBonawitz2019,JKonecny2016,TLi2020,HBMcMahan2017,VSmith2017}, leaving significant energy efficiency 5.4 times and model convergence 4.2 times co-optimization opportunity on the table (Figure~\ref{FL-motivation-oracle}).

{\bf System Heterogeneity and Stochastic Runtime Variance:} At the edge, there are over two thousand unique System-on-Chips (SoCs) with different compute resources, including CPUs, graphic processing units (GPUs), and digital signal processors (DSPs), in more than ten thousand different smart devices~\cite{CJWu2019,YGKim2020}. The high degree of system heterogeneity introduces varying, potentially large, performance gaps across smartphones in FL. In addition, mobile execution is stochastic by nature~\cite{BGaudette2016,BGaudette2019,CJWu2019}, due to co-running application interference and network stability. All of these lead to the straggler problem---training time of each aggregation round is gated by the slowest participant smartphone.
To mitigate the straggler problem, several previous works built on top of FedAvg by excluding stragglers from each round~\cite{HBMcMahan2017} or by allowing partial updates from the straggler~\cite{TLi2020}. However, these approaches sacrifice accuracy. 

{\bf Data Heterogeneity:} Varying characteristics of training data per participating device introduce additional challenge to efficient FL execution~\cite{ZChai2019,XLi2020}. To guarantee model convergence, it is important to ensure that training data is independently and identically distributed (IID) across the participating devices~\cite{ZChai2019,HGuan2020,FYan2015}. For example, if a model is developed to classify images into 10 distinct label categories, the data samples are IID if each individual participant device has independent data, representing all 10 categories~\cite{CWang2021}. However, in a realistic environment, the training data samples on each device are usually based on the behavior and/or preference of users. Thus, local training samples of any particular user will not be representative of the population, deferring convergence~\cite{XLi2020,HBMcMahan2017}. To mitigate data heterogeneity, previous approaches proposed to exclude the non-IID devices~\cite{YChen2019,YChen2019_2}, to use a warm up model~\cite{YZhao2018}, or to share data across a subset of participant devices~\cite{MDuan2021,ALi2020}. However, none has considered both data \textit{and} system heterogeneity with runtime variance.

Furthermore, there has been very little work on energy efficiency optimization for FL. Most prior work assume that FL is only activated when smartphones are plugged into wall-power, due to the significant energy consumption of model training~\cite{YChen2019_2,XQiu2020,ZXu2019,CWang2021,YZhan2020}. Unfortunately, this has limited the practicality of FL, resulting in longer model convergence time and degraded model accuracy~\cite{WYBLim2020,CWang2021}. Energy efficient FL could enable on-device training anytime, with better model quality and user experience.

To tackle challenges from realistic execution environment, this paper proposes a learning-based energy optimization framework---\textit{AutoFL}---that selects \textit{K} participants as well as execution targets to guarantee model quality, while maximizing energy efficiency of individual participants (or the cluster of all participating smartphones in aggregate) for FL. The optimization is performed by considering the presence of system and data heterogeneity and runtime variance. Since the optimal decision varies with NN characteristics, FL global parameters, profiles of participating devices, distributions of local training samples, and stochastic runtime variance, the design space is massive and infeasible to enumerate. Thus, we design a reinforcement learning technique. For each aggregation round, AutoFL observes NN characteristics, FL global parameters, and system profiles of devices (including interference intensity, network stability, and data distributions). It then selects the participant devices for the round and, at the same time, determines the execution target for each participant, to maximize energy efficiency while guaranteeing the training accuracy requirements. The result of the decision is measured and fed back to AutoFL, allowing it to continuously learn and predict the optimal action for the subsequent rounds. 

AutoFL is implemented and run on the centralized, model aggregation server. We evaluate our proposed design using 200 mobile systems composed of three major categories of mobile systems, representative of high, medium, and low performance levels. The real-system evaluation demonstrates that, compared to the baseline random selection setting, AutoFL improves the energy efficiency of the individual smartphones by an average of 4.7 times and the overall energy efficiency for the entire cluster by 5.2 times, maintaining the training accuracy.
The key contributions of this work are as follows:
\begin{itemize}
\item We present an in-depth performance and energy efficiency characterization for FL by considering realistic edge-cloud execution environment. The results show that the optimal participant selection and resource allocation in FL can vary significantly with neural network characteristics, the varying degree of data and system heterogeneity, and the stochastic nature of mobile execution (Section~\ref{sec:motivation}).
\item We propose a learning-based FL energy optimization framework, called AutoFL. AutoFL identifies near-optimal participant selection and resource allocation at runtime, enabling heterogeneity-aware energy efficient federated learning (Section~\ref{sec:design}).
\item To demonstrate the feasibility and practicality, we design, implement, and evaluate AutoFL for a variety of FL use cases in the edge-cloud environment. Real-system experiments show that AutoFL improves energy efficiency of individual participant devices as well as the cluster of all participating devices by an average of 4.7x and 5.2x, respectively, while also satisfying the accuracy requirement (Section~\ref{sec:result}).
\end{itemize}

\section{Background}
\label{sec:background}
\vspace{-0.2cm}

\begin{figure}[t]
    \centering
    \includegraphics[width=\linewidth]{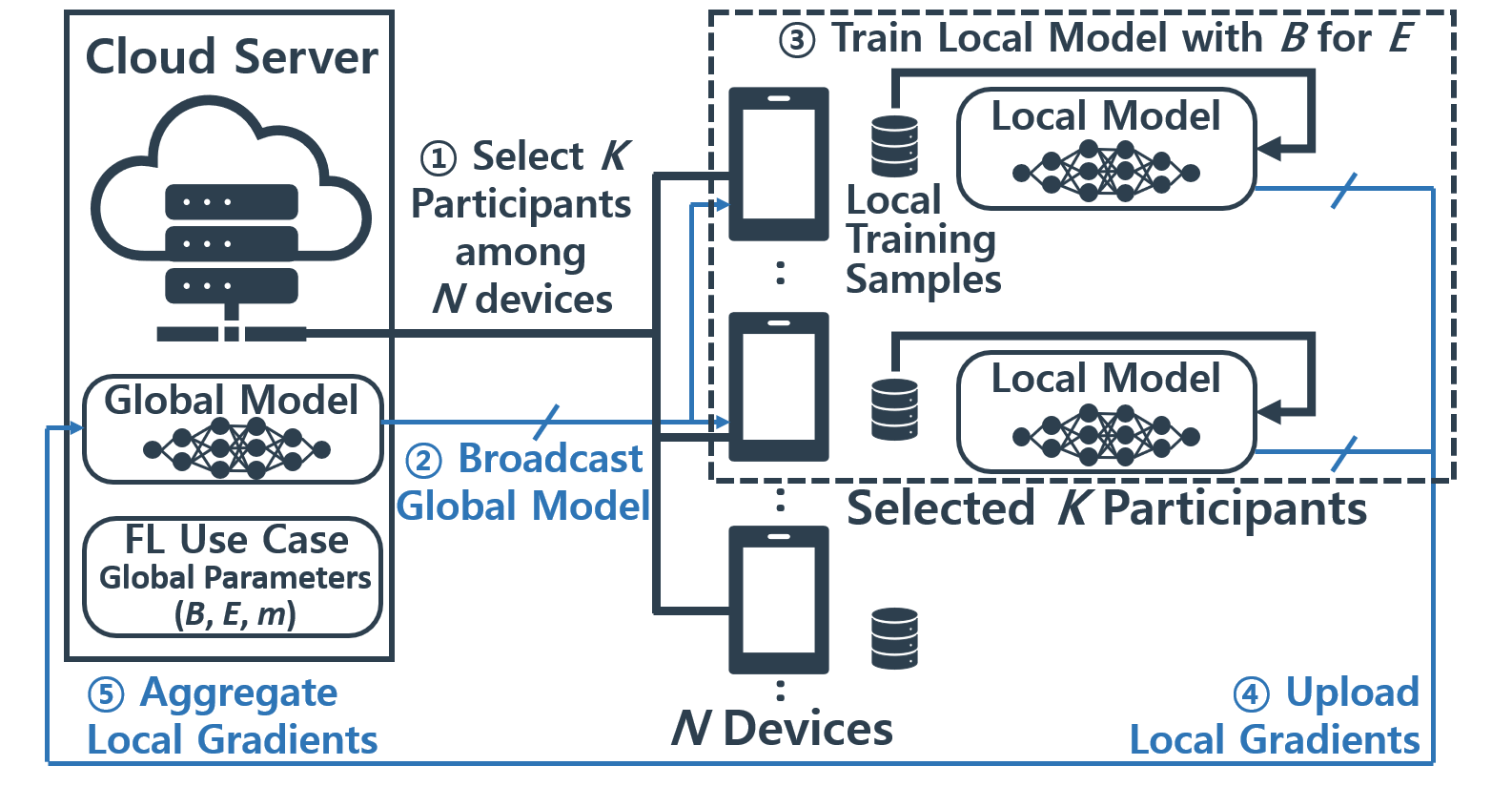}
    \vspace{-0.3cm}
    \caption{Overview for Federated Learning.}
    \label{fig:background1}
\end{figure}

\subsection{Federated Learning}
\label{sec:background1}
To improve data privacy for ML training, Federated Learning (FL) is introduced by allowing local devices at the edge, such as smartphones, to collaboratively train a shared ML model while keeping all user data locally on the device~\cite{KBonawitz2019,RCGeyer2017,SItahara2020,JKonecny2016,ALi2020,WYBLim2020,HBMcMahan2017,ZTao2018,CWang2021,YZhan2020,JZhang2020}.
Figure~\ref{fig:background1} depicts the overall system architecture for the federated learning baseline~\cite{KBonawitz2019,JKonecny2016,HBMcMahan2017}. There are two entities in the FL system---an \textit{aggregation server} as the model owner and a collection of local devices (data owners). Given \textit{N} local devices for FL, the server first initializes a global deep learning model and its global parameters by specifying the number of local epochs \textit{E} for training, the local training minibatch size \textit{B}, and the number of participant devices \textit{K}. (\textit{B}, \textit{E}, \textit{K}) is determined by the FL-based services~\cite{KBonawitz2019,HBMcMahan2017}.

In each aggregation round, the server selects \textit{K} participant devices among the \textit{N} devices (Step \textcircled{\small{1}}) and broadcasts the global model to the selected devices (Step \textcircled{\small{2}}). Each participant independently trains the model by using the local data samples with the batch size of \textit{B} for \textit{E} epochs (Step \textcircled{\small{3}}). Once the local training step finishes, the computed model gradients are sent back to the server (Step \textcircled{\small{4}}). The server then aggregates the local gradients and calculates the average of the local gradients~\cite{HBMcMahan2017} to update the global model (Step \textcircled{\small{5}}). The steps are repeated until a desirable accuracy is achieved. 

\subsection{Consideration for Realistic Execution Environment}
\label{sec:background2}

\begin{figure}[t]
    \centering
    \includegraphics[width=0.95\linewidth]{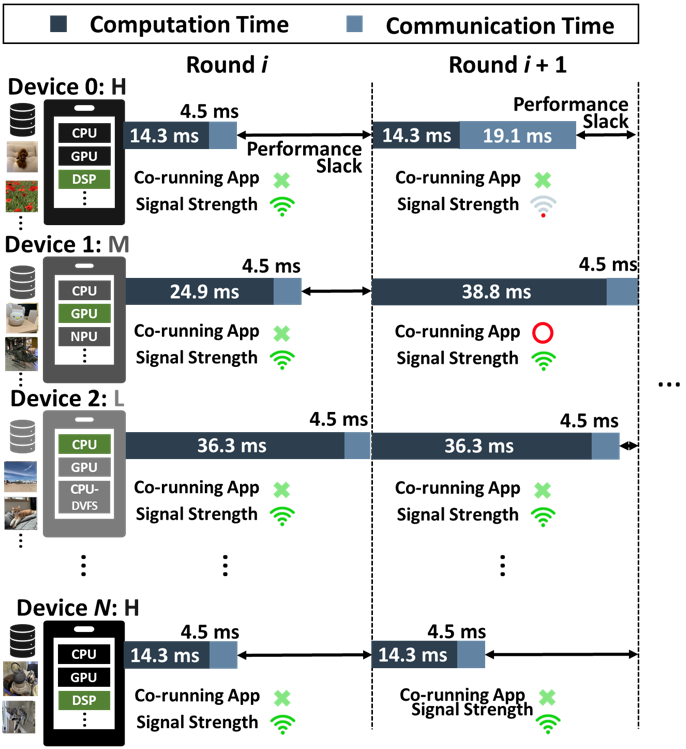}
    \vspace{-0.25cm}
    \caption{Optimization space of FL is large, considering the scale of decentralized training, system heterogeneity, data heterogeneity, and sources of runtime variance.}
    \label{FL-Design-Space}
\end{figure}


System heterogeneity, runtime variance, and data heterogeneity form a massive optimization space for FL.
Figure~\ref{FL-Design-Space} illustrates FL execution in a realistic environment. In this example, a cluster of two hundred devices participate in FL. Depending on the performance level of an individual device (i.e., high-end, mid-end, and low-end smartphones) and the availability of co-processors, such as GPUs, DSPs, or neural processing units (NPUs), the training time performance varies. High degree of system heterogeneity introduces large performance gaps across the devices, leading to the straggler problem~\cite{TLi2020,WYBLim2020,HBMcMahan2017,CWang2021,YZhan2020}. 

In addition, stochastic runtime variance can exacerbate the straggler problem. Depending on the amount of on-device interference and the execution conditions, such as ambient temperatures and network signal strength, the execution time performance of each individual participant---the training time per round (\textit{Computation Time}) and model gradient aggregation time (\textit{Communication Time})---is highly dynamic. Finally, not all participant devices possess IID training samples. Heterogeneous data across devices can significantly deteriorate FL model convergence and quality.

\section{Motivation}
\label{sec:motivation}
This section presents system characterization results for FL.
We examine the design space covering three important axes --- energy efficiency, convergence time, and accuracy. 

\subsection{Impact of FL Global Parameters and NN Characteristics}
\label{sec:motivation1}

\begin{figure}[t]
    \centering
    \includegraphics[width=\linewidth]{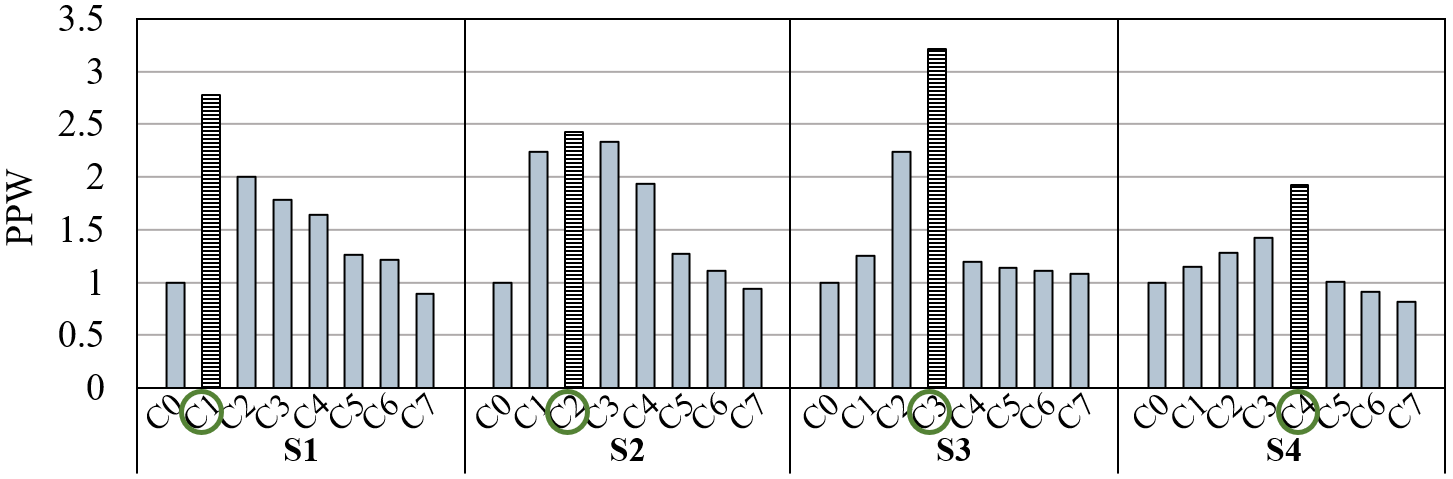}
    \vspace{-0.3cm}
    \caption{Depending on global parameters of FL use cases and NN model resource needs, the optimal clusters of {\it K} participating devices are {\tt C1}, {\tt C2}, {\tt C3}, and {\tt C4} across the four different global parameter settings, respectively. The detailed description for the settings and the clusters is in Table~\ref{table:global_parameter} and Table~\ref{table:cluster} of Section~\ref{sec:methodology}, respectively. Striped bars indicate the optimal cluster.  
    }
    \label{FL-global-parameter}
\end{figure}

The optimal cluster of participating devices depends on the global parameters of FL and the resource requirement of specific NN models. From the system's perspective, the global parameters determine the amount of computations performed on each individual device.
Figure~\ref{FL-global-parameter} compares the achieved energy efficiency under four different FL global parameter settings ({\tt S1} to {\tt S4} defined in Table~\ref{table:global_parameter} of Section~\ref{sec:methodology2}) for training the CNN model with MNIST dataset (CNN-MNIST) over eight different combinations of participant devices ({\tt C0} to {\tt C7} defined in Table~\ref{table:cluster}. The optimal device cluster changes from {\tt C1} to {\tt C2}, {\tt C3} and {\tt C4} when the global parameter setting changes from {\tt S1} to {\tt S2}, {\tt S3}, and {\tt S4}, respectively.

When the number of computations assigned to each device is large (i.e., {\tt S1}), including more high-end devices is beneficial as they exhibit 1.7x and 2.5x better training time, compared to mid-end and low-end devices, respectively, due to powerful CPUs and co-processors along with larger size of cache and memory. On the other hand, when the number of computations assigned to each device decreases (i.e., from {\tt S1} to {\tt S2} and {\tt S3}), including mid-end and low-end devices along with the high-end devices results in better energy efficiency since their lower power consumption (35.7\% and 46.4\% compared to high-end devices respectively) amortizes the performance gap reduced due to lower computation and memory requirements. If \textit{K} is decreased (i.e., from {\tt S3} to {\tt S4}), reducing the number of high-end devices is beneficial as the devices can stay idle during the round --- though mid-end devices have longer training-time-per-round than high-end devices, similar to {\tt S3}, the mid-end devices have better energy efficiency in this case. 

When we use the LSTM model with Shakespeare dataset (denoted as LSTM-Shakespeare), the optimal device cluster is {\tt C3}, {\tt C4}, {\tt C5}, and {\tt C5}, respectively, as compared to CNN-MNIST's {\tt C1}, {\tt C2}, {\tt C3}, {\tt C4} over {\tt S1}--{\tt S4}. In the case of CNN-MNIST, due to the compute-intensive CONV and FC layers, high-end devices with more powerful mobile SoCs show better performance and energy efficiency, compared to the mid- and low-end devices. 
On the other hand, for LSTM-Shakespeare, the energy efficiency of mid- and low-end devices is comparable to that of high-end devices. This is because the performance difference among the devices gets smaller (from 2.1x to 1.5x, on average) due to the memory operations, so that the low power consumption of the mid- and low-end devices amortizes their performance loss.

\subsection{Impact of Runtime Variance}
\label{sec:motivation2}

\begin{figure}[t]
    \centering
    \includegraphics[width=\linewidth]{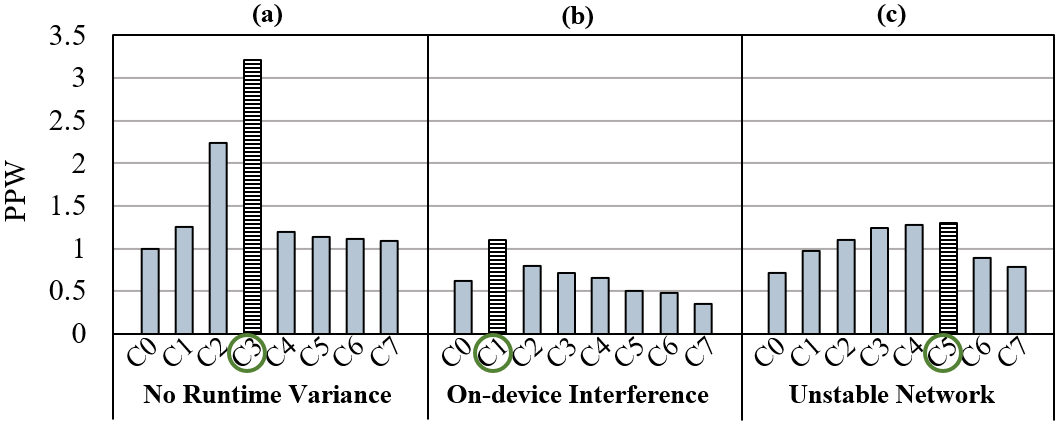}
    \caption{With runtime variance from various sources, the optimal cluster of {\it K} participating devices shifts from {\tt C3} to {\tt C1} and {\tt C5}. PPW is normalized to {\tt C0} with no runtime variance.}
    \label{FL-runtime-variance}
    \vspace{-0.2cm}
\end{figure}

The optimal cluster of participants also significantly varies along with the runtime variance. Figure~\ref{FL-runtime-variance}(a) compares the achieved energy efficiency in the absence of on-device interference and with stable network signal strength. In such an ideal execution environment, the most energy efficient cluster is {\tt C3}, balancing the trade-off between the training-time-per-round and power consumption of different categories of devices --- {\tt C3} achieves 3.2x higher energy efficiency than the baseline {\tt C0}. In the presence of on-device interference, the optimal cluster becomes {\tt C1} (Figure~\ref{FL-runtime-variance}(b)), whereas when the network signal strength is weak, the optimal cluster switches to {\tt C5} (Figure~\ref{FL-runtime-variance}(c)).

Intuitively, in the presence of on-device interference, it is more beneficial to select high-end devices to participate in FL--- those devices have a high computation and memory capabilities~\cite{YGKim2020}, so that they show 2.0x and 3.1x better performance compared to mid-end and low-end devices, respectively. On the other hand, when the network signal strength is poor, the communication time and energy on each device is significantly increased~\cite{NDing2013,YGKim2019} (4.3x, on average). In this case, the impact of performance gap among different categories of devices decreases along with the decreased portion of computation time. For this reason, including low-power devices is beneficial in terms of energy efficiency due to lower computation and communication power consumption.

\subsection{Impact of Data Heterogeneity}
\label{sec:motivation3}

Participant device selection strategies that ignore data heterogeneity lead to sub-optimal FL execution. Figure~\ref{FL-data-heterogeneity}(a) shows the convergence patterns for CNN-MNIST over varying degrees of data heterogeneity---the x-axis shows the consecutive FL rounds and the y-axis shows the model accuracy.
Here, {\tt Non-IID (M\%)} means M\% of \textit{K} participant devices have non-IID data where a proportion of the samples of each data class is distributed following Dirichlet distribution with 0.1 of concentration parameter~\cite{ZChai2019,QLi2021,XLi2020,TLin2020,PKairouz2019}, while the rest of devices have all the data classes independently --- the smaller the value of the concentration parameter is, the more each data class is concentrated to one device.

\begin{figure}[t]
    \centering
    \includegraphics[width=\linewidth]{Figures/Fig.5.png}
    \vspace{-0.3cm}
    \caption{(a) Model quality and (b) energy efficiency of FL changes with varying levels of data heterogeneity.}
    \label{FL-data-heterogeneity}
    \vspace{-0.3cm}
\end{figure}

\begin{figure*}[t]
    \centering
    \includegraphics[width=\linewidth]{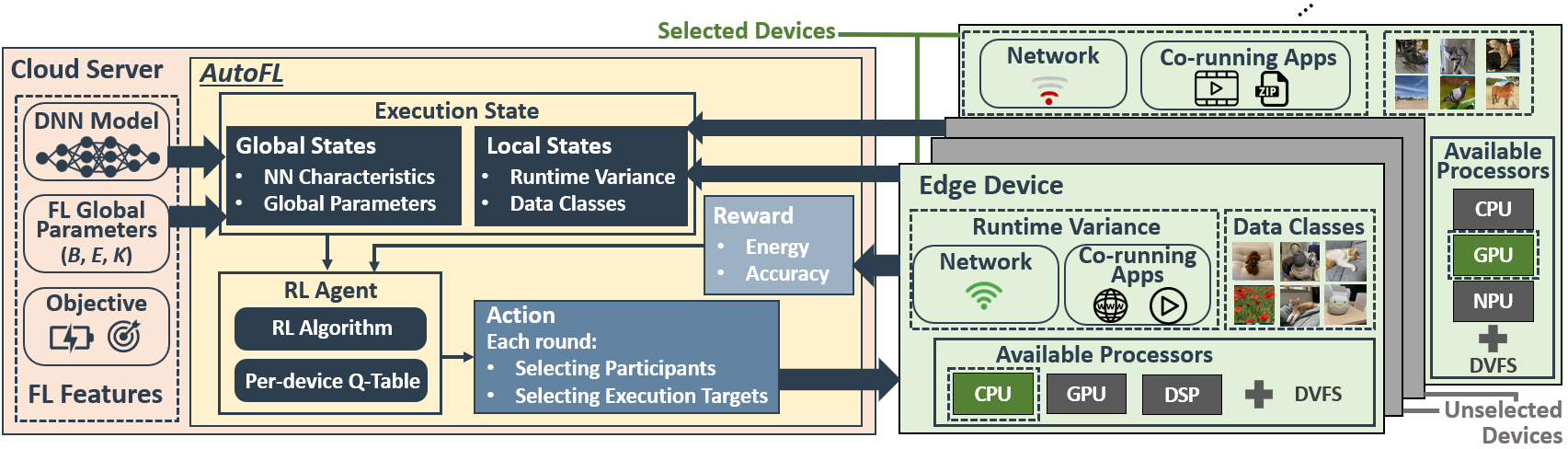}
    \vspace{-0.3cm}
    \caption{Design Overview for AutoFL.}
    \label{fig:design}
\end{figure*}

Data heterogeneity can significantly affect model convergence---when devices with non-IID data participate in FL, the convergence time is significantly increased compared to the ideal IID scenario. The increased convergence time eventually deteriorates FL energy efficiency. Figure~\ref{FL-data-heterogeneity}(b) illustrates the large (>85\%) energy efficiency gap between the ideal device selection scenario and the sub-optimal selection scenarios with non-IID data.

\section{AutoFL}
\label{sec:design}

To capture stochastic runtime variance in the presence of system and data heterogeneity, we propose an adaptive prediction mechanism based on reinforcement learning\footnote{We exploit RL instead of other statistical methods, such as Gaussian Process, since RL has the following advantages: (1) faster training and inference due to lower complexity~\cite{FBerns2021,SPagani2020}, (2) higher sample efficiency (i.e., the amount of experiments to reach a certain level of accuracy)~\cite{SMohanty2021,SCKao2020,DYarats2019}, and (3) higher prediction accuracy under the stochastic variance~\cite{YGKim2020,SPagani2020}.}, called AutoFL. In general, an RL agent learns a policy to select the best action for a given state with accumulated rewards~\cite{SPagani2020}. 
In the context of FL, given a NN and the corresponding global parameters, AutoFL learns to select an optimal cluster of participating devices and an energy-efficient execution target in individual devices for each aggregation round. 

Figure~\ref{fig:design} provides the design overview of AutoFL. 
During each FL aggregation round, AutoFL observes the global configurations of FL, including target NN and the global parameters. In addition, it collects the execution states of participant devices, including their resource usage and network stability\footnote{AutoFL relies on the resource usage and network bandwidth information collected by de-facto FL protocol~\cite{KBonawitz2019} --- the protocol collects such information to ensure model training robustness. Note, to avoid the leak of system usage information, it is possible to run training of per-device Q-table locally, without sharing the information with the cloud server at the expense of increased training cost (336.2 $\mu$s in low-end device including communication cost.).}, and the number of data classes each device has. Based on the information, AutoFL identifies the current execution state.
For the identified state, AutoFL selects participant devices that are expected to maximize the energy efficiency of FL, while satisfying the accuracy requirement.
AutoFL also determines the execution target for each selected device to additionally improve the local energy efficiency.
The selections are based on per-device lookup tables (i.e., Q-tables) that contain the accumulated rewards of previous selections.
After the gradient updates are aggregated in the server, AutoFL measures the results (i.e., training time, energy consumption, and test accuracy) to calculate the reward---how the selected action improves global as well as local energy efficiency and accuracy.
Finally, AutoFL updates the per-device Q-table with the calculated reward.
To solve system optimization using RL, there are three important design requirements.

\textbf{High Prediction Accuracy:} High prediction accuracy is essential to the success of an RL-based approach. To handle the dynamic execution environment of FL, it is important to model the core components---state, action, and reward---in a realistic environment directly. We define the components in accordance with our observations of a realistic FL execution environment (Section~\ref{sec:design1}).
In addition to the core components, avoiding local optima is also important. The fine balance between exploitation versus exploration is at the heart of RL~\cite{EE-Dar2006,DEKoulouriotis2008}.
If an RL agent always exploits an action with the temporary highest reward, it can get stuck in a local optima. On the other hand, if it keeps exploring all possible actions, convergence of rewards in RL may take too long. To tackle this challenge, we employ the epsilon-greedy algorithm. This algorithm is one of the commonly-adopted algorithms~\cite{YGKim2020,SKMandal2020,RNishtala2017,SPagani2020} due to its effectiveness and simplicity (Section~\ref{sec:design2}) --- it achieves comparable prediction accuracy to other complex algorithms, such as Exp3, Softmax, UCB, and Thompson Sampling, with lower overhead~\cite{DCortes2018,IUmami2021}.

\textbf{Low Training and Inference Overhead:} To minimize the timing and energy overhead for on-device RL, AutoFL expedites the training of the RL model by enabling devices within the same performance category to share the learned results --- in a realistic environment, each user can experience different degree of the data heterogeneity and runtime variance, and thus sharing the learned results across the devices complements one another. We present the training time overhead reduction of this approach in Section~\ref{sec:result3}.

The inference latency of per-device RL models determines the decision making performance of AutoFL. Thus, among the various RL implementation choices, e.g., Q-learning~\cite{YChoi2019} and deep RL~\cite{VMnih2015}, Q-learning is most suitable to this work --- it achieves low training and inference latency with look-up tables, while deep RL usually exhibits longer latency because of forward and backward propagation of DNNs~\cite{SPagani2020}. Hence, in this paper, we use Q-learning for AutoFL.

\begin{scriptsize}
\begin{table*}[t]
  \caption{State Features for AutoFL.}
  \vspace{0.2cm}
  \centering
  \begin{tabular}{|l|l|l|l|}
    \hline
    \multicolumn{2}{|l|}{\textbf{State}} & \textbf{Description} & \textbf{Discrete Values} \\
    \hline
        \multirow{3}{1.4cm}{NN-related Features} & $S_{CONV}$ & \# of CONV layers & Small (<10), medium (<20), large (<30), larger (>=40) \\ \cline{2-4}
        & $S_{FC}$ & \# of FC layers & Small (<10), large (>=10) \\ \cline{2-4}
        & $S_{RC}$ & \# of RC layers & Small (<5), medium (<10), large (>=10) \\  \hline
        \multirow{3}{1.4cm}{Global Parameters} & $S_{B}$ & Batch size & Small (<8), medium (<32), large (>=32) \\ \cline{2-4}
        & $S_{E}$ & \# of local epochs & Small (<5), medium (<10), large (>=10) \\ \cline{2-4}
        & $S_{K}$ & \# of participant devices & Small (<10), medium (<50), large (>=50) \\ \hline
        \multirow{3}{1.4cm}{Runtime Variance} & $S_{Co\_CPU}$ & CPU utilization of 
        co-running apps & None (0\%), small (<25\%), medium (<75\%), large (<=100\%)\\ \cline{2-4}
        & $S_{Co\_MEM}$ & Memory usage of co-running apps & None (0\%), small (<25\%), medium (<75\%), large (<=100\%)\\ \cline{2-4}
        & $S_{Network}$ & Network bandwidth & Regular (>40Mbps), bad (<=40Mbps) \\ \hline
        Data Classes& $S_{Data}$ & \# of data classes for this round & Small (<25\%), medium (<100\%), large (=100\%)\\ \hline
  \end{tabular}
  \vspace{-0.2cm}
  \label{table:States}
\end{table*}
\end{scriptsize}

\textbf{Scalability:} As energy efficient FL can enable many more devices to participate in FL, the scalability to a large number of devices is crucial. In order to scale to a large number of participating devices, AutoFL can exploit a shared Q-table for devices within the same performance category --- additional clustering algorithm can be used along with the AutoFL for binding similar category of devices. By updating the shared Q-table instead of all the per-device Q tables, AutoFL can deal with the large number of devices, at the expense of a small prediction accuracy loss (see details in Section~\ref{sec:result3}).

\vspace{-0.2cm}
\subsection{AutoFL Reinforcement Learning Design}
\label{sec:design1}

We define the core RL components---\textit{State}, \textit{Action}, and \textit{Reward}---to formulate the optimization space for AutoFL.

\textbf{State:} Based on the observations presented in Section~\ref{sec:motivation}, we identify states that are critical to energy-efficient FL execution. Table~\ref{table:States} summarizes the states.

First, the energy efficiency of devices participating in FL highly depends on NNs and the given global parameters. In order to model the impact of NN characteristics and global parameters, we identify states with layer types that are deeply correlated with the energy efficiency and performance of on-device training execution. We test the correlation strength between each layer type and energy efficiency by calculating the squared correlation coefficient ($\rho^{2}$)~\cite{YZhu2013}. We find convolution layers (CONV), fully-connected layers (FC), and recurrent layers (RC) impact energy efficiency differently due to their respective compute- and/or memory-intensive natures. Thus, we identify $S_{CONV}$, $S_{FC}$, and $S_{RC}$ to represent the number of CONV, FC, and RC layers in a NN, respectively. We also identify  $S_{B}$, $S_{E}$, and $S_{K}$ to represent global parameters of batch size, the number of local epochs, and the number of participant devices, respectively.

Energy efficiency of participating devices are highly influenced by the runtime variance---namely, on-device interference and network stability. To model on-device interference, we identify per-device states of $S_{Co\_CPU}$ and $S_{Co\_MEM}$ to represent CPU utilization and memory usage of co-running applications, respectively. We also model per-device network stability with $S_{Network}$ to represent the network bandwidth of the respective wireless network (e.g., Wi-Fi, LTE, and 5G). 
In addition, data heterogeneity also has a significant impact on the convergence time and energy efficiency of FL. Therefore, to model the impact of data heterogeneity on the FL efficiency, we identify $S_{Data}$ which stands for the number of data classes that each device has for an aggregation round.

When a feature has a continuous value, it is difficult to define the state in a discrete manner for the lookup table of Q-learning~\cite{YChoi2019,YGKim2020,RNishtala2017}. To convert the continuous features into discrete values, we applied the DBSCAN clustering algorithm to each feature~\cite{YChoi2019,YGKim2020}---DBSCAN determines the optimal number of clusters for the given data. The last column of Table~\ref{table:States} summarizes the discrete values.

\textbf{Action:} Actions in reinforcement learning represents the tunable control knobs of a system. In the context of FL, we define the actions in two levels. At the global level, we define the selection of participant devices as an action. For each selected participant device, we define the selection of on-device execution targets available for training execution, such as CPUs, GPUs, or DSP, as another action. The execution targets are augmented to include CPU dynamic voltage and frequency scaling (DVFS) settings, to exploit the performance slack caused by stragglers for further energy saving.

\textbf{Reward:} In RL, rewards track the optimization objective of the system. To represent the main optimization axes, we encode three rewards: $R_{energy\_local}$, $R_{energy\_global}$, and $R_{accuracy}$. $R_{energy\_local}$ is the estimated energy consumption of each individual participant device and $R_{energy\_global}$ is the estimated energy consumption of the cluster of all participating devices. $R_{accuracy}$ represents the test accuracy of the NN.

We estimate $R_{energy\_local}$ and $R_{energy\_global}$ as follows. For each selected participant device, we first calculate the computation energy, $E_{comp}$. When the CPU is selected as the execution target, $E_{comp}$ is calculated using a utilization-based CPU power model \cite{DBrooks2000,RJoseph2001,YGKim2020,LZhang2010} as in (\ref{eq: CPU energy}), where $E^{i}_{core}$ is the power consumed by the \textit{i}th core, $t^{f}_{busy}$ and $t_{idle}$ are the time spent in the busy state at frequency \textit{f} and that in the idle state, respectively, and $P^{f}_{busy}$ and $P_{idle}$ are the power consumed during $t^{f}_{busy}$ at \textit{f} and that during $t_{idle}$, respectively.
\vspace{-0.2cm}
\begin{equation}
    \label{eq: CPU energy}
    \vspace{-0.1cm}
    \begin{aligned}
        E_{comp} &= \sum_{i}{E_{core}^{i}}, \\
        E_{core} &= \sum_{f}(P_{busy}^{f} \times t_{busy}^{f}) + P_{idle} \times t_{idle} \\
    \end{aligned}
    \vspace{-0.2cm}
\end{equation}
Similarly, if GPU is selected as the execution target, $E_{comp}$ is calculated using the GPU power model \cite{NDing2017} as in (\ref{eq: GPU energy}). Note that $t^{f}_{busy}$ and $t_{idle}$ for CPU/GPU are obtained from \textit{procfs} and \textit{sysfs} in the Linux kernel \cite{YChoi2019}, while $P^{f}_{busy}$ and $P_{idle}$ for CPU/GPU are obtained by power measurement of CPU/GPU at each frequency in the busy state and idle state, respectively\footnote{Although we only present the energy estimation of CPU and GPU in this paper due to the limited programmability of on-device training, similar practice can also be used for other co-processors, such as DSPs and NPUs~\cite{MHan2019,SWang2020}.}. Those values are obtained for representative categories of edge devices (i.e., high-end, mid-end, and low-end devices) and stored in a look-up table of AutoFL.

\vspace{-0.3cm}
\begin{equation}
    \label{eq: GPU energy}
    E_{comp} = \sum_{f}(P_{busy}^{f} \times t_{busy}^{f}) + P_{idle} \times t_{idle}
    \vspace{-0.2cm}
\end{equation}
After calculating the computation energy, we calculate the communication energy, $E_{comm}$, for each selected participant using the signal strength-based energy model~\cite{YGKim2019} as in (\ref{eq:communication}), where $t_{TX}$ is the latency measured while transmitting the gradient updates, and $P^{S}_{TX}$ is power consumed by a wireless network interface during $t_{TX}$ at signal strength \textit{S}. Note $P^{S}_{TX}$ is obtained by measuring power consumption of wireless network interfaces at each signal strength, transmitting data.
\vspace{-0.1cm}
\begin{equation}
    \label{eq:communication}
    \begin{aligned}
        E_{comm} = &P_{TX}^{S} \times t_{TX} 
    \end{aligned}
\vspace{-0.1cm}
\end{equation}
We also calculate the idle energy, $E_{idle}$, for non-selected devices, as in (\ref{eq:idle}), where $t_{round}$ is the time spent during the round for the training.
\vspace{-0.1cm}
\begin{equation}
    \label{eq:idle}
    \begin{aligned}
        E_{idle} = & P_{idle} \times t_{round}
    \end{aligned}
\vspace{-0.1cm}
\end{equation}
Based on the estimated energy values, $R_{energy\_local}$ is calculated for each device, as in (\ref{eq:Renergy_local}), where $S_{t}$ represents a set of selected participants.
\vspace{-0.2cm}
\begin{equation}
    \label{eq:Renergy_local}
    \vspace{-0.1cm}
    \begin{aligned}
        &if \;\; device \;\; \subset \;\; S_{t} \\
        &\qquad R_{energy\_local} = E_{comp} + E_{comm}\\
        &else \\
        &\qquad R_{energy\_local} = E_{idle}
    \end{aligned}
\end{equation}
In addition, $R_{energy\_global}$ is calculated for a cluster of all \textit{N} participating devices, as in (\ref{eq:Renergy_global}), based on the $R_{energy\_local}$.
\vspace{-0.2cm}
\begin{equation}
    \label{eq:Renergy_global}
    \vspace{-0.1cm}
    \begin{aligned}
        R_{energy\_global} = & \sum_{i}^{N}R_{energy\_local}
    \end{aligned}
    \vspace{-0.2cm}
\end{equation}
Since the energy estimation is based on the measured latency, its mean absolute percentage error is 7.3\%---low enough to identify the optimal participants and execution targets.

To ensure AutoFL selects participants and execution targets that maximize energy efficiency while satisfying the accuracy targets, the reward \textit{R} is calculated as in (\ref{eq:reward})\footnote{We include the energy consumption as reward to model the impact of selections on the global and local energy efficiency. We include accuracy to model the impact of selections on model quality.}, 
where $R_{accuracy\_prev}$ is the test accuracy of the training NN from the previous round. $\alpha$ and $\beta$ are the weights for the accuracy and the amount of accuracy improvement which is directly related to the convergence speed, respectively. 
\begin{equation}
    \label{eq:reward}
    \begin{aligned}
        &if \;\;\; R_{accuracy} \; - \; R_{accuracy\_prev} \;\; <= \;\; 0,\\
        &\qquad R = R_{accuracy} - 100\\
        &else \\
        &\qquad R = - R_{energy\_global} - R_{energy\_local} \\ 
        &\qquad \;\;\;\;\;\;\;\; + \alpha R_{accuracy} + \beta (R_{accuracy} - R_{accuracy\_prev})\\
    \end{aligned}
\end{equation}
If the selected action fails to improve the accuracy from the previous round, the reward is $R_{accuracy}$ -- 100 (i.e., how much the accuracy is far from 100\%) to avoid choosing the action for the next inference. Otherwise, the reward is calculated for each device based on the global energy, local energy, accuracy, and the amount of accuracy improvement.


\subsection{AutoFL Implementation Detail}
\label{sec:design2}
AutoFL is built based on Q-learning. To strike a balance between exploitation and exploration in RL, AutoFL employs the epsilon-greedy algorithm with a uniformly random action, based on a pre-specified exploration probability. For the rest, AutoFL chooses an action with the highest reward.

\begin{scriptsize}
\begin{algorithm}[t]
\caption{Training the Q-Learning Model}
\label{alg: training}
\textbf{Variable:} $S_{global}$, $S_{local}$, \textit{A} \\
\hspace*{\algorithmicindent}\hspace*{\algorithmicindent} $S_{global}$ is the global state \\
\hspace*{\algorithmicindent}\hspace*{\algorithmicindent} $S_{local}$ is the local state \\
\hspace*{\algorithmicindent}\hspace*{\algorithmicindent} \textit{A} is the action (execution target) \\
\textbf{Constants:} \textit{$\gamma$, $\mu$, $\epsilon$} \\
\hspace*{\algorithmicindent}\hspace*{\algorithmicindent} \textit{$\gamma$} is the learning rate \\
\hspace*{\algorithmicindent}\hspace*{\algorithmicindent} \textit{$\mu$} is the discount factor \\
\hspace*{\algorithmicindent}\hspace*{\algorithmicindent} \textit{$\epsilon$} is the exploration probability \\
\textbf{Initialize} $Q(S_{global}, S_{local}, A)$ as random values \\
\textbf{Repeat} (whenever an aggregation round begins): \\
\hspace*{\algorithmicindent}\hspace*{\algorithmicindent} Observe global state and store in $S_{global}$ \\
\hspace*{\algorithmicindent}\hspace*{\algorithmicindent} Observe local state for each device and store in $S_{local}$ \\
\hspace*{\algorithmicindent}\hspace*{\algorithmicindent} \textbf{if} rand() < \textit{$\epsilon$} \textbf{then} \\
\hspace*{\algorithmicindent}\hspace*{\algorithmicindent}\hspace*{\algorithmicindent} Choose \textit{K} participants randomly \\
\hspace*{\algorithmicindent}\hspace*{\algorithmicindent}\hspace*{\algorithmicindent} Choose action \textit{A} randomly for selected participants \\
\hspace*{\algorithmicindent}\hspace*{\algorithmicindent}\textbf{else} \\
\hspace*{\algorithmicindent}\hspace*{\algorithmicindent}\hspace*{\algorithmicindent} Sort devices by $Q(S_{global}, S_{local}, A)$ \\
\hspace*{\algorithmicindent}\hspace*{\algorithmicindent}\hspace*{\algorithmicindent} Choose at most top \textit{K} participants \\
\hspace*{\algorithmicindent}\hspace*{\algorithmicindent}\hspace*{\algorithmicindent} Choose action \textit{A} with the largest $Q(S_{global}, S_{local}, A)$ \\
\hspace*{\algorithmicindent}\hspace*{\algorithmicindent} Run training on a target defined by \textit{A} in each device\\
\hspace*{\algorithmicindent}\hspace*{\algorithmicindent} (when local training and aggregation ends) \\
\hspace*{\algorithmicindent}\hspace*{\algorithmicindent} Estimate $R_{energy\_global}$, $R_{energy\_local}$, and obtain $R_{accuracy}$ \\
\hspace*{\algorithmicindent}\hspace*{\algorithmicindent} Calculate reward \textit{R} \\
\hspace*{\algorithmicindent}\hspace*{\algorithmicindent} Observe new global state $S'_{global}$ \\
\hspace*{\algorithmicindent}\hspace*{\algorithmicindent} Observe new local state $S'_{local}$ \\
\hspace*{\algorithmicindent}\hspace*{\algorithmicindent} Sort devices by $Q(S'_{global}, S'_{local}, A')$ \\
\hspace*{\algorithmicindent}\hspace*{\algorithmicindent} Choose at most top \textit{K} participants \\
\hspace*{\algorithmicindent}\hspace*{\algorithmicindent} Choose action \textit{A'} with the largest $Q(S'_{global}, S'_{local}, A')$ \\
\hspace*{\algorithmicindent}\hspace*{\algorithmicindent} $Q(S_{global}, S_{local}, A)$ $\leftarrow$ $Q(S_{global}, S_{local}, A)$ \\
\hspace*{\algorithmicindent}\hspace*{\algorithmicindent} \;\;\;\;\;\;\;\;\;\;\;\;\;\;\;\;\;\;\;\;\;\;\;\;\;\;\;\;\;\;\; +  $\gamma$[\textit{R} + $\mu$ $Q(S'_{global}, S'_{local}, A')$ \\
\hspace*{\algorithmicindent}\hspace*{\algorithmicindent} \;\;\;\;\;\;\;\;\;\;\;\;\;\;\;\;\;\;\;\;\;\;\;\;\;\;\;\;\;\;\; -- $Q(S_{global}, S_{local}, A)$ \\
\hspace*{\algorithmicindent}\hspace*{\algorithmicindent} \textit{S} $\leftarrow$ \textit{S'} 
\end{algorithm}
\end{scriptsize}

In Q-learning, the value function $Q(S_{global}, S_{local}, A)$ takes the global state $S_{global}$, local state $S_{local}$, and action \textit{A} as parameters in the form of a lookup table (Q-table). Algorithm 1 shows the detailed algorithm for training the per-device Q-table. At the beginning, AutoFL initializes the Q-tables with random values. At runtime, AutoFL observes $S_{global}$ and $S_{local}$ for each aggregation round, by checking the NN characteristics, runtime variance, and data heterogeneity. It evaluates a random value compared with $\epsilon$\footnote{Note that we use 0.1 for $\epsilon$ based on our sensitivity analysis.}. If the random value is smaller than $\epsilon$, AutoFL selects participants randomly and determines \textit{A} for exploration. Otherwise, it sorts the devices by $Q(S_{global}, S_{local}, A)$ and selects the top \textit{K} devices.

Next, AutoFL chooses \textit{A} with the largest $Q(S_{global}, S_{local}, A)$ for each selected participant. After the local training and the aggregation ends, AutoFL estimates $R_{energy\_local}$ and $R_{energy\_global}$ as explained in Section~\ref{sec:design1}. In addition, it obtains $R_{accuracy}$ and $R_{accuracy\_prev}$. Based on these values, AutoFL calculates the reward \textit{R} as in (\ref{eq:reward}) of Section~\ref{sec:design1}. After that, AutoFL observes the new states and chooses the corresponding participants and execution targets with the $Q(S'_{global}, S'_{local}, A')$. It then updates the $Q(S_{global}, S_{local}, A)$ based on the equation in Algorithm 1. In the equation, $\gamma$ and $\mu$ are hyperparmeters that represent the learning rate and discount factor, respectively. We set $\gamma$ and $\mu$ based on the sensitivity evaluation. Hyperparameter tuning is described in Section~\ref{sec:methodology3}.

After learning is completed, i.e., the largest $Q(S_{global}, S_{local}, A)$ value for each $S_{global}$ and $S_{local}$ is converged, the collection of per-device Q-tables are used to select participants and the corresponding \textit{A} which maximizes $Q(S_{global}, S_{local}, A)$ for the observed $S_{global}$ and $S_{local}$. Note, among the devices which have the same $Q(S_{global}, S_{local}, A)$, AutoFL randomly selects participants to avoid biased selection~\cite{TLi2020,TLi2020_2}.
\section{Experimental Methodology}
\label{sec:methodology}

\subsection{System Measurement Infrastructure}
\label{sec:methodology1}
We set up an edge-cloud FL system that consists of 200 mobile devices (\textit{N} = 200) and one model aggregation server. Similar FL system infrastructures have been used in a number of prior works~\cite{MDuan2021,TLi2020,HBMcMahan2017,JKonecny2016}.
We emulate the performance of FL execution by using Amazon EC2 instances~\cite{Amazon2} that provide the same theoretical GFLOP performance as the three representative categories of smartphones---high-end (H), mid-end (M), and low-end (L) devices. Table~\ref{table:device} summarizes the system profiles. Among the 200 instances, there are 30 H, 70 M, and 100 L devices, representative of in-the-field system performance distribution~\cite{CJWu2019}. 

For model aggregation, we connect the aforementioned systems to a high-performance Amazon EC2 system instance, c5d.24xlarge, which has a theoretical performance of 448 GFLOPS and is equipped with 32GB of RAM. 
We perform power measurement directly using an external Monsoon Power Meter~\cite{Monsoon} for the three smartphones during on-device training (implemented with DL4j~\cite{Dl4j}): Mi8Pro~\cite{Huawei}, Galaxy S10e~\cite{Samsung}, and Moto X Force~\cite{Motorola} (Table~\ref{table:Devices}). Similar power measurement methodologies are used in prior works~\cite{DPandiyan2013,DPandiyan2014,DShingari2015}. 

Based on the measured performance and power consumption, we evaluate the energy efficiency of participant devices in FL. 
To characterize the FL energy efficiency with various clusters of participant devices, we compare the energy efficiency of various clusters of devices (Table~\ref{table:cluster}) in Section~\ref{sec:motivation}. Based on the characterization results, we build AutoFL as described in Section~\ref{sec:design}, and implement it upon the state-of-the-art FedAvg algorithm~\cite{JKonecny2016,HBMcMahan2017} using PyTorch~\cite{PyTorch}. 

To evaluate the effectiveness of AutoFL, we compare it with five other design points:
\begin{compactitem}
\item the {\tt FedAvg-Random} baseline~\cite{HBMcMahan2017} where {\it K} participants are determined randomly, 
\item {\tt Power} where {\it K} participants are determined by minimizing power draw (i.e., {\tt C7} in Table 4), 
\item {\tt Performance} where {\it K} participants are selected to achieve best execution time performance (i.e., {\tt C1} in Table 4), 
\item {\tt $O_{participant}$} where the optimal cluster of {\it K} participants is determined by considering heterogeneity and runtime variance, and 
\item {\tt $O_{FL}$} that considers available on-device co-processors for energy efficiency improvement over {\tt $O_{participant}$}. 
\end{compactitem}
We also compare AutoFL with two closely-related prior works: FedNova~\cite{JWang2020} and FEDL~\cite{CTDinh2021}.

\begin{scriptsize}
\begin{table}[t]
  \caption{Amazon EC2 Instance Specification.}
  \vspace{0.2cm}
  \centering
  \begin{tabular}{|c|c|c|c|}
    \hline
    \multirow{2}{0.8cm}{\textbf{Level}} & \multirow{2}{1.2cm}{\textbf{Instance}} & \textbf{Performance} & \textbf{RAM} \\ 
    & & \textbf{(GFLOPS)} & \textbf{(GB)} \\ \hline
    H & m4.large & 153.6 & 8 \\ \hline
    M & t3a.medium & 80 & 4 \\ \hline
    L & t2.small & 52.8 & 2 \\ \hline
  \end{tabular}
  \label{table:device}
\end{table}
\end{scriptsize}

\begin{scriptsize}
\begin{table}[t]
  \caption{Mobile Device Specification.}
  \vspace{0.2cm}
  \centering
  \begin{tabular}{|c|c|c|}
    \hline
    \textbf{Device}& \textbf{CPU}& \textbf{GPU} \\
    \hline
    \multirow{3}{1.2cm}{Mi8Pro (H)} & Cortex A75 (2.8GHz) & Adreno 630 (0.7GHz) \\ 
    & 23 V-F steps & 7 V-F steps \\
    & 5.5 W & 2.8 W \\
    \hline
    \multirow{3}{1.2cm}{Galaxy S10e (M)} & Mongoose (2.7GHz) & Mali-G76 (0.7GHz) \\ 
    & 21 V-F steps & 9 V-F steps \\
    & 5.6 W & 2.4 W \\
    \hline
    \multirow{3}{1.2cm}{Moto X Force (L)} & Cortex A57 (1.9GHz) & Adreno 430 (0.6GHz) \\ 
    & 15 V-F steps & 6 V-F steps \\
    & 3.6 W & 2.0 W \\
    \hline
  \end{tabular}
  \label{table:Devices}
\end{table}
\end{scriptsize}

\vspace{-0.1cm}
\subsection{Workloads and Execution Scenarios}
\label{sec:methodology2}

{\bf Workloads:} We evaluate AutoFL using two commonly-used FL workloads: (1) training the CNN model with the MNIST dataset ({\bf CNN-MNIST}) for image classification~\cite{YLeCun1998,YLeCun1998_2,JTSpringenberg2015} and (2) training the LSTM model with the Shakespeare dataset ({\bf LSTM-Shakespeare}) for the next character prediction~\cite{HBMcMahan2017,JKonecny2016}. The workloads are widely used and representative of state-of-the-art FL use cases~\cite{MDuan2021,TLi2020,HBMcMahan2017,JKonecny2016}.
In addition, we complement CNN-MNIST and LSTM-Shakespeare with an additional workload: (3) training the MobileNet model with the ImageNet dataset ({\bf MobileNet-ImageNet}) for image classification~\cite{JDeng09,AHoward2017}.
Table~\ref{table:global_parameter} summarizes the value range of the global parameters we consider in this work. 
Note, once the global parameters are determined for an FL use case, the values stay fixed until the model convergence~\cite{KBonawitz2019,JKonecny2016}.

\begin{scriptsize}
\begin{table}[t]
  \caption{Cluster of Devices Used for Characterization.}
  \vspace{0.2cm}
  \centering
  \begin{tabular}{|c|c|c|c|c|}
    \hline
    \textbf{Cluster} & \textbf{H} & \textbf{M} & \textbf{L} & \textbf{Policy} \\ \hline
    C0 & - & - & - & {\tt FedAvg-Random} (Baseline) \\ \hline 
    C1 & 20 & 0 & 0 & {\tt Performance}\\ \hline
    C2 & 15 & 5 & 0 & \\ \hline
    C3 & 10 & 5 & 5 & \\ \hline
    C4 & 5 & 10 & 5 & \\ \hline
    C5 & 5 & 5 & 10 & \\ \hline
    C6 & 0 & 5 & 15 & \\ \hline
    C7 & 0 & 0 & 20 & {\tt Power}\\ \hline
  \end{tabular}
  \label{table:cluster}
\end{table}
\end{scriptsize}

\begin{scriptsize}
\begin{table}[t]
  \caption{Global Parameter Settings.}
  \centering
  \begin{tabular}{|c|c|c|c|}
    \hline
    \textbf{Setting} & \textbf{{\it B}} & \textbf{{\it E}} & \textbf{{\it K}} \\ \hline
    S1 & 32 & 10 & 20 \\ \hline
    S2 & 32 & 5 & 20 \\ \hline
    S3 & 16 & 5 & 20 \\ \hline
    S4 & 16 & 5 & 10 \\ \hline
  \end{tabular}
  \label{table:global_parameter}
\end{table}
\end{scriptsize}


{\bf Runtime Variance:} To emulate realistic on-device interference, we initiate a synthetic co-running application on a random subset of devices, mimicking the effect of a real-world application, i.e., web browsing~\cite{YHuang2014,MJu2016,DShingari2015,DShingari2018,DPandiyan2013}. The synthetic application generates CPU and memory utilization patterns following those of web browsing. In addition, since the real-world network variability is typically modeled by a Gaussian distribution~\cite{NDing2013}, we emulate the random network bandwidth with a Gaussian distribution by adjusting the network delay.

{\bf Data Distribution:} We emulate different levels of data heterogeneity by distributing the total training dataset in four different ways~\cite{ZChai2019,XLi2020}: Ideal IID, Non-IID (50\%), Non-IID (75\%), and Non-IID (100\%).
In case of Ideal IID, all the data classes are evenly distributed to the cluster of total devices.
On the other hand, in case of Non-IID ({\tt M\%}), {\tt M\%} of total devices have non-IID data while the rest of devices have IID samples of all the data classes. For non-IID devices, we distributed each data class randomly following Dirichlet distribution with 0.1 of concentration parameters~\cite{ZChai2019,QLi2021,XLi2020,TLin2020,PKairouz2019} --- the smaller the value of the concentration parameter is, the more each data class is concentrated to one device.

\subsection{AutoFL Design Specification}
\label{sec:methodology3}

\textbf{Actions:} We determine the 2-level actions for AutoFL. The first-level action determines a cluster of participant devices (Section~\ref{sec:design1}) whereas the second-level action determines an execution target for the FL execution.
Since the energy efficiency of local devices can be further improved via DVFS when stragglers are present, we identify V/F steps available in the FL system~\cite{Amazon2} as the augmented second-level action.
Note, we measure the power consumption of different mobile devices with varying frequency steps, in order to accurately model the energy efficiency of the FL execution. 

\textbf{Hyperparameters:} There are two hyperparameters in the FL system: the learning rate and discount factor. To determine them, we evaluate three values of 0.1, 0.5, and 0.9 for each hyperparameter~\cite{YChoi2019,YGKim2020}. We observe that the learning rate of 0.9 shows 20.1\% and 32.5\% better prediction accuracy, compared to that of 0.5 and 0.1, respectively, meaning the more portion of the reward is added to the Q values, the better AutoFL works. This is because AutoFL needs to adapt to the runtime variance and data heterogeneity in the limited aggregation rounds.
On the other hand, we observe that the discount factor of 0.1 shows 20.1\% and 53.4\% better prediction accuracy compared to that of 0.5 and 0.9, respectively, meaning the less portion of reward value for the next state is added to that for the current state, the better AutoFL works. This is because the consecutive states have a weak relationship due to the stochastic nature, so that giving less weight to the reward in the near future improves the efficiency of AutoFL. Thus, in our evaluation, we use 0.9 for the learning rate and 0.1 for the discount factor.
\section {Evaluation Results and Analysis}
\label{sec:result}

\vspace{-0.1cm}
\subsection{Result Overview}
\label{sec:result1}
\vspace{-0.1cm}

\begin{figure}[t]
    \centering
    \includegraphics[width=\linewidth]{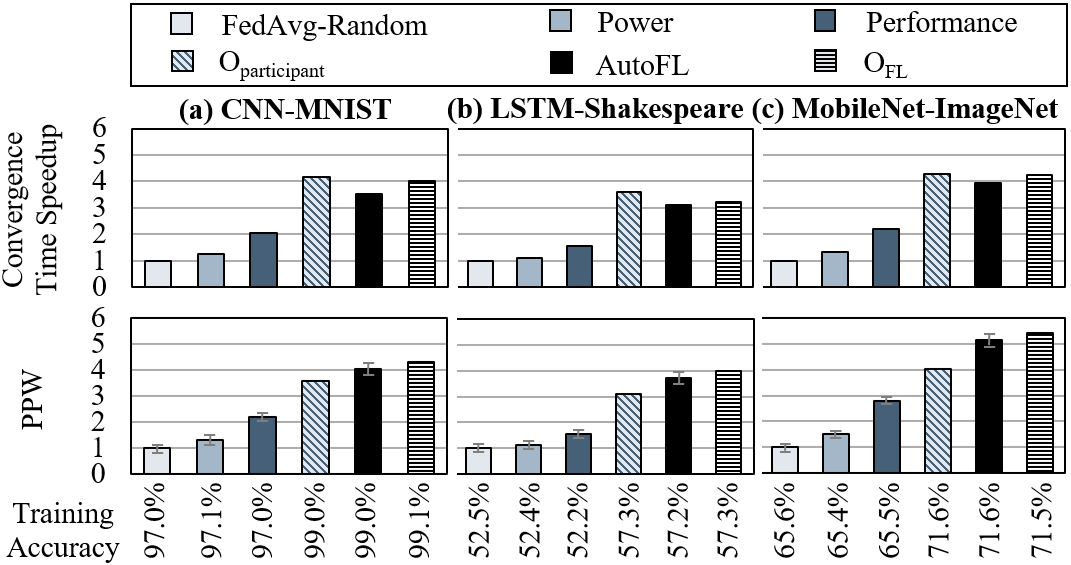}
    \vspace{-0.3cm}
    \caption{AutoFL improves convergence time and energy efficiency of FL, while also increasing model quality. It achieves 4.0x, 3.7x, and 5.1x higher energy efficiency over the baseline FedAvg-Random for CNN-MNIST, LSTM-Shakespear, and MobileNet-ImageNet, respectively.
    }
    \label{fig:overall}
\end{figure}

Compared with the baseline settings of {\tt FedAvg-Random}, {\tt Power}, and {\tt Performance}, AutoFL significantly improves the average FL energy efficiency of CNN-MNIST, LSTM-Shakespeare, and MobileNet-ImageNet by 4.3x, 3.2x, and 2.0x, respectively. In addition, AutoFL shows better training accuracy. Figure~\ref{fig:overall} compares the energy efficiency in performance-per-watt (PPW), the convergence time, and training accuracy for the respective FL use cases where PPW and the convergence time improvement are normalized to the {\tt FedAvg-Random} baseline.

The energy efficiency gains of AutoFL come from two major sources. First, AutoFL can accurately identify optimal participants among a wide variety for each FL use case, reducing the performance slack from the stragglers. As a result, it improves the training time per round by an average of 3.5x, 2.9x, and 1.8x, over {\tt FedAvg-Random}, {\tt Power}, and {\tt Performance}, respectively. This leads to faster convergence time. Second, for the individual participants, AutoFL identifies more energy efficient execution targets. By doing so, the energy efficiency is improved further by an average of 19.8\% over {\tt $O_{participant}$}. Compared to $O_{participant}$, AutoFL and {\tt $O_{FL}$} experience slightly higher convergence time. This is because AutoFL leverages the remaining performance slack by considering alternative on-device execution targets and DVFS settings despite the slight increase in computation time.

In the case of CNN-MNIST and MobileNet-ImageNet, compute-intensive CONV and FC layers are dominant. In this case, performance-oriented selection, i.e., {\tt Performance}, shows much higher energy efficiency, compared to power-oriented selection, i.e., {\tt Power}, due to the higher computation and memory capabilities of high-end devices. On the other hand, in the case of LSTM-Shakespeare, compute- and memory-intensive RC layers are dominant. In this case, the difference between the performance-oriented selection, i.e., {\tt Performance}, and the power-efficient selection, i.e., {\tt Power}, decreases.  Nevertheless, since the baseline settings do not consider the NN characteristics explicitly in the participant device selection process, AutoFL's achieved energy efficiency outweighs that of other design points. 

\subsection{Adaptability and Accuracy Analysis}
\label{sec:result2}

\begin{figure}[t]
    \centering
    \includegraphics[width=\linewidth]{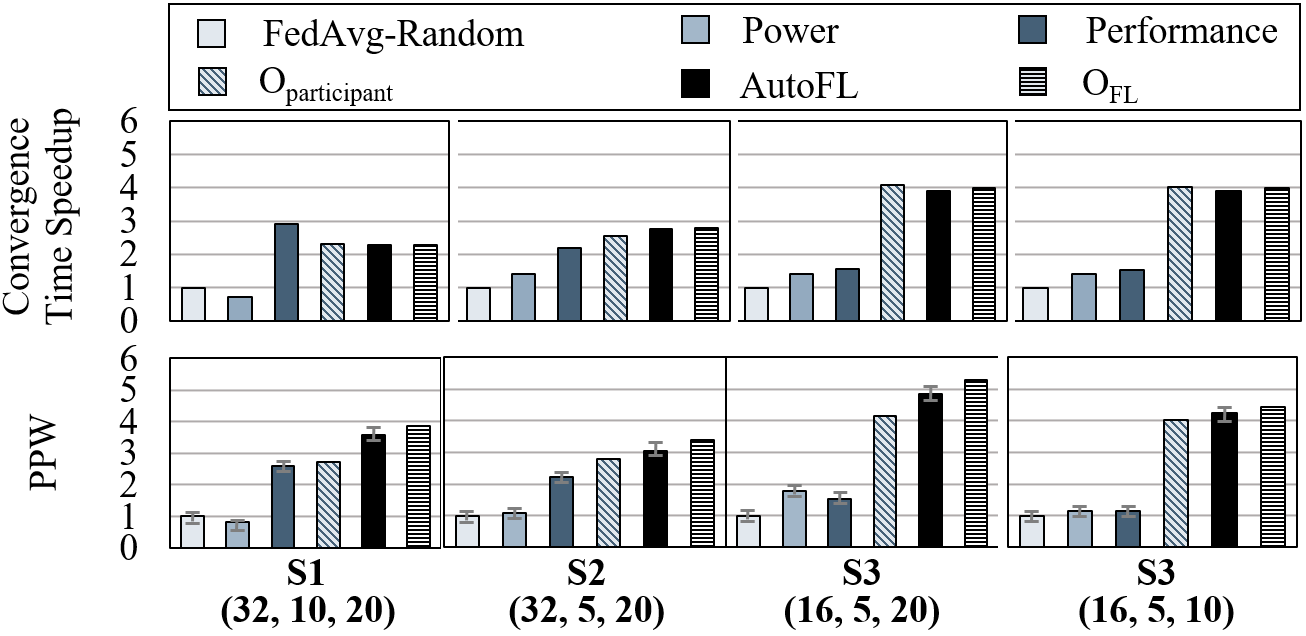}
    \vspace{-0.3cm}
    \caption{Across the (\textit{B}, \textit{E}, \textit{K}) global parameter settings of {\tt S1}--{\tt S4}, AutoFL achieves better training time performance and higher energy efficiency consistently.}
    \label{fig:global}
\end{figure}

{\bf Adaptability to Global Parameters:} AutoFL significantly improves the energy efficiency and convergence time for various combinations of global parameters.  Figure~\ref{fig:global} shows the average energy efficiency and convergence time of CNN-MNIST across four different global parameter settings---$S1$ to $S4$ (Table 4 in Section~\ref{sec:methodology2}).
Although the optimal cluster of participant devices varies along with the global parameters (as we observed in Section~\ref{sec:motivation}), AutoFL accurately predicts the optimal cluster of participant devices regardless of global parameters. Hence, AutoFL always outweighs the baseline settings of {\tt FedAvg-Random}, {\tt Performance}, and {\tt Power} in terms of energy efficiency and convergence time. In addition, since AutoFL also accurately predicts the optimal execution targets for individual devices, it achieves 15.9\% better energy efficiency, compared to {\tt $O_{participant}$}.

\begin{figure}[t]
    \centering
    \includegraphics[width=\linewidth]{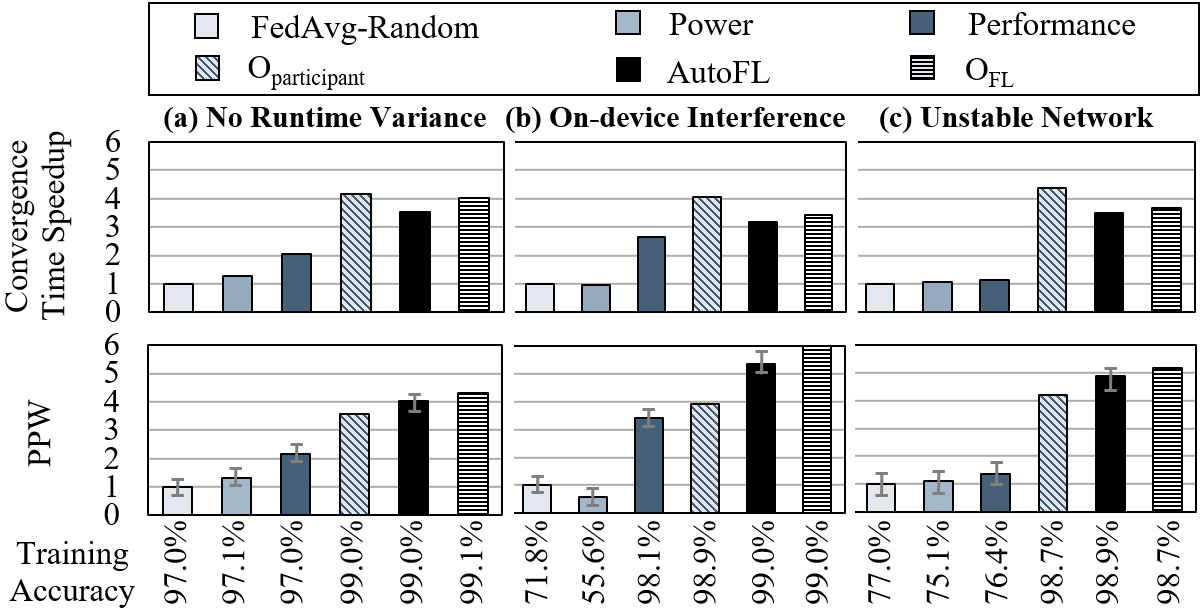}
    \vspace{-0.3cm}
    \caption{
    In the presence of runtime variance, AutoFL can consistently and significantly improve the time-to-convergence and energy efficiency for FL under different execution environments.
    }
    \label{fig:realistic environment}
\end{figure}

{\bf Adaptability to Stochastic Variance:} AutoFL can improve the energy efficiency and convergence time in the presence of stochastic on-device interference and network variance, independently. Figure~\ref{fig:realistic environment} shows the PPW, convergence time, and training accuracy of CNN-MNIST, (a) when there is no on-device intererence, and when there is (b) on-device interference from co-running applications or (c) network variance. Even in the presence of runtime variance, AutoFL improves the average energy efficiency by 5.1x, 6.9x, and 2.6x, compared to {\tt FedAvg-Random}, {\tt Power}, and {\tt Performance}. Note other NNs also show similar result trends.

In the presence of runtime variance, the training time per round of the baseline settings significantly increases because of the increased on-device computation time or communication time. Even worse, since FedAvg algorithm excludes the severe stragglers from the round, the convergence time as well as the training accuracy is additionally degraded. On the other hand, AutoFL accurately selects the optimal cluster of participant devices even in the presence of runtime variance, mitigating the straggler problem. By doing so, it improves the convergence time by 3.4x, 3.3x, and 2.3x, compared to {\tt FedAvg-Random}, {\tt Power}, and {\tt Performance}, respectively. Additionally, AutoFL also exploits the increased performance gap from the stragglers, improving 26.3\% more energy efficiency compared to {\tt $O_{participant}$} at the cost of training time per round. As a result, AutoFL achieves almost similar energy efficiency, convergence time, and training accuracy with {\tt $O_{FL}$}. 

\begin{figure}[t]
    \centering
    \includegraphics[width=\linewidth]{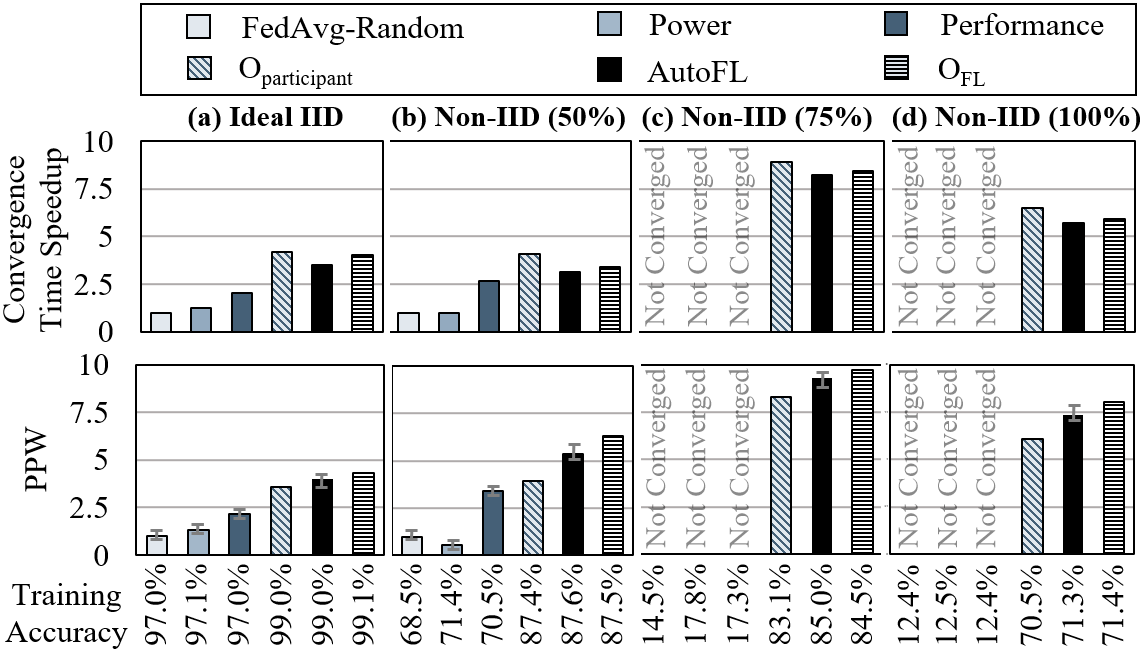}
    \vspace{-0.3cm}
    \caption{By explicitly taking into account data heterogeneity in the selection of {\it K} devices, AutoFL achieves 4.0x, 5.5x, 9.3x, and 7.3x higher energy efficiency over the baseline FedAvg-Random for the four data distribution scenarios: (a)--(d).}
    \label{fig:data}
\end{figure}

\textbf{Adaptability to Data Heterogeneity:} In the presence of data heterogeneity, compared to {\tt FedAvg-Random}, {\tt Power}, and {\tt Performance}, AutoFL significantly improves the energy efficiency by 7.4x, 5.5x, and 4.3x, respectively. It also shows much better convergence time and training accuracy. Figure~\ref{fig:data} illustrates the energy efficiency, convergence time, and training accuracy of CNN-MNIST. Each column shows the varying level of data heterogeneity: (a) Ideal IID, (b) Non-IID (50\%), (c) Non-IID (75\%), (d) Non-IID (100\%). Note other NNs also show similar result trends.

When there exist non-IID participants, the baseline settings (i.e., {\tt FedAvg-Random}, {\tt Power}, and {\tt Performance}) that do not consider data heterogeneity experience sub-optimal energy efficiency, convergence time, and training accuracy. This is because naively including non-IID participants can significantly deteriorate model convergence — in the case of Non-IID (75\%) and Non-IID (100\%), CNN-MNIST does not even converge with the baseline settings in 1000 rounds (Figure~\ref{fig:data}(c) and Figure~\ref{fig:data}(d)).
In contrast, AutoFL \textit{learns the impact of data heterogeneity on the convergence time and energy efficiency dynamically} and \textit{adapts to the different level of data heterogeneity across the devices}. Therefore, it achieves near-optimal energy efficiency, convergence time, and model quality even in the presence of data heterogeneity.

\begin{figure}[t]
    \centering
    \includegraphics[width=0.95\linewidth]{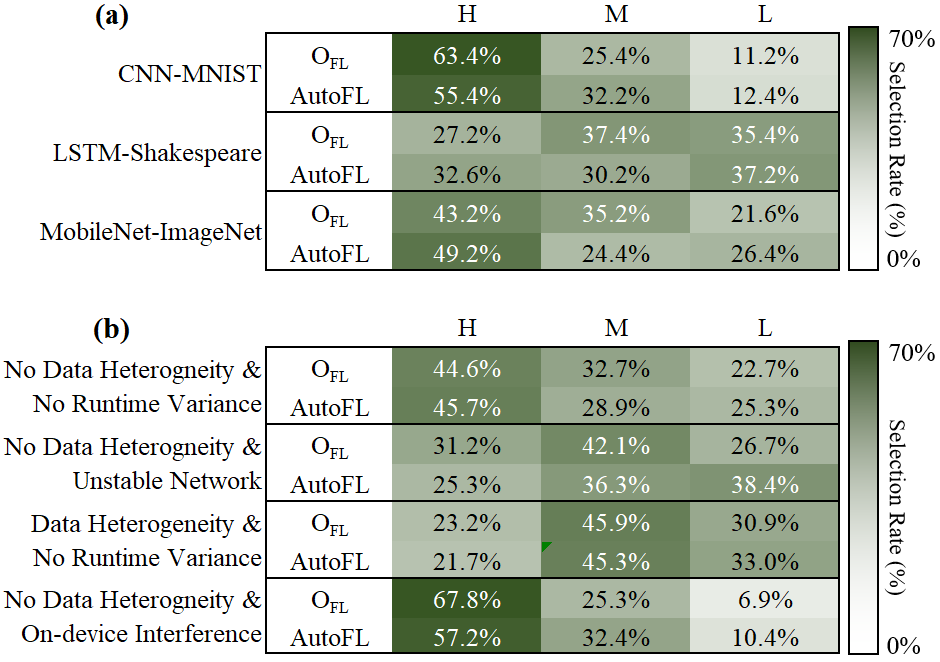}
    \caption{AutoFL can track the decisions from the optimal policy ($O_{FL}$) accurately.}
    \label{fig:prediction}
\end{figure}

\textbf{Prediction Accuracy:} AutoFL accurately selects the optimal cluster of participants in varying data heterogeneity and runtime variance for the given NNs.  Figure~\ref{fig:prediction} shows how AutoFL and $O_{FL}$ make the participant selection on three different categories of devices. For participant selection. AutoFL achieves 93.9\% of average prediction accuracy.

AutoFL accurately selects the optimal cluster of participants for different NNs. In Figure~\ref{fig:prediction}(a), the optimal cluster of participant devices significantly vary depending on the NN characteristics. For example, {\tt $O_{FL}$} includes more high-end devices CNN-MNIST and MobileNet-ImageNet, whereas it includes more mid-end and low-end devices for LSTM-Shakespeare. AutoFL accurately captures those trends, achieving 94.2\% of average accuracy.

AutoFL also accurately adapts to the data heterogeneity and runtime variance. As shown in Figure~\ref{fig:prediction}(b), even in the presence of runtime variance and data heterogeneity, AutoFL accurately makes the optimal participant selection, achieving 93.7\% prediction accuracy on average.

AutoFL also accurately selects the optimal execution targets in individual participant device. When there is no runtime variance, CPU rather than GPU shows better energy efficiency, because of compute- and memory-intensive nature of training workloads. On the other hand, when there exists on-device interference, the optimal execution target usually shifts from CPU to GPU, since the CPU performance is significantly degraded due to 1) the competition for CPU time slice and cache, and 2) frequent thermal throttling. In case of unstable network, CPU and GPU show similar energy efficiency as FL becomes communication bound. AutoFL accurately captures such impact of runtime variance on optimal excution targets, achieving 92.9\% average accuracy. Hence, as shown in Figure~\ref{fig:overall}, \ref{fig:global}, \ref{fig:realistic environment}, and \ref{fig:data}, AutoFL substantially improves energy efficiency, compared to {\tt $O_{participant}$} which does not select the optimal execution target in each participant.

\subsection{Comparison with Prior Work}
\label{sec:result_prior}

\begin{figure}[t]
    \centering
    \includegraphics[width=0.95\linewidth]{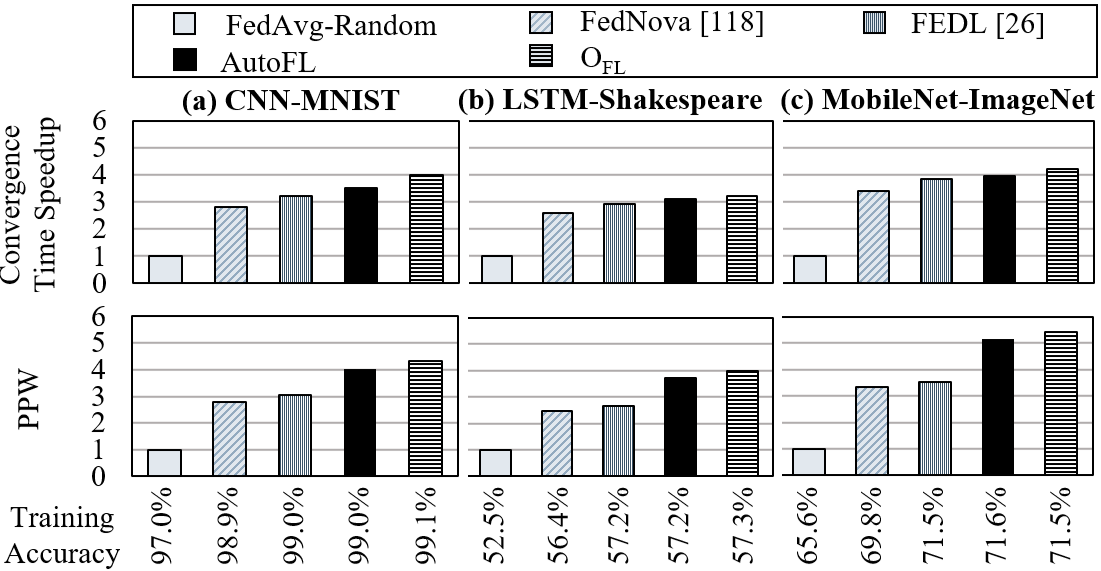}
    \caption{As compared with the prior works: FedNova~\cite{JWang2020} and FEDL~\cite{CTDinh2021}, AutoFL achieves better convergence time and higher energy efficiency for FL.}
    \label{fig:prior}
\end{figure}

\begin{figure}[t]
    \centering
    \includegraphics[width=0.95\linewidth]{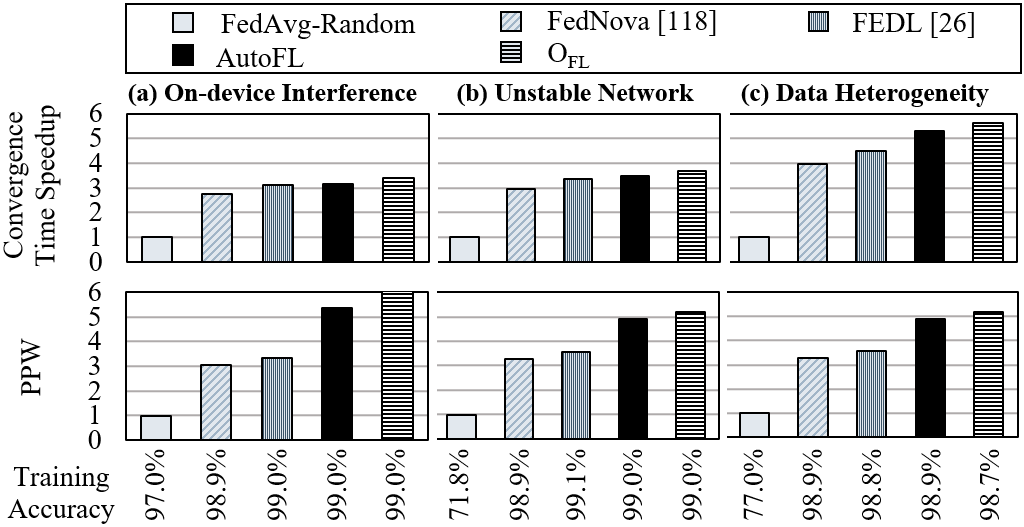}
    \caption{AutoFL outperforms both FedNova~\cite{JWang2020} and FEDL~\cite{CTDinh2021}, even in the presence of runtime variance (a)(b) and data heterogeneity (c).}
    \label{fig:prior2}
\end{figure}

We compare AutoFL with two closely-related prior works: FedNova~\cite{JWang2020} and FEDL~\cite{CTDinh2021}. FedNova normalizes gradient updates from stragglers or non-IID devices to those from ideal devices while FEDL lets each client device to approximately adjust gradient updates based on the global weights. Both FedNova and FEDL allow partial updates from stragglers but implement random participant selections. Furthermore, neither work considers exploiting other available execution targets to accelerate FL performance or energy efficiency. On average, compared with FedNova and FEDL, AutoFL achieves 49.8\% and 39.3\% higher energy efficiency, respectively (Figure~\ref{fig:prior}).
In the presence of stochastic variance, FedNova and FEDL improve the execution time performance and PPW over the baseline, as expected. Similarly, AutoFL can further increase PPW by 62.7\% and 48.8\% over FedNova and FEDL, respectively (Figure~\ref{fig:prior2}).

Compared with the baseline, FedNova and FEDL are robust to data heterogeneity by giving less weights to gradient updates from non-IID devices. Nonetheless, including non-IID users can degrade model convergence, lowering time-to-convergence and energy efficiency. In contrast, AutoFL achieves near-optimal energy efficiency, convergence time, and model quality, even in the presence of data heterogeneity.

\subsection{Overhead Analysis}
\label{sec:result3}

\begin{figure}[t]
    \centering
    \includegraphics[width=\linewidth]{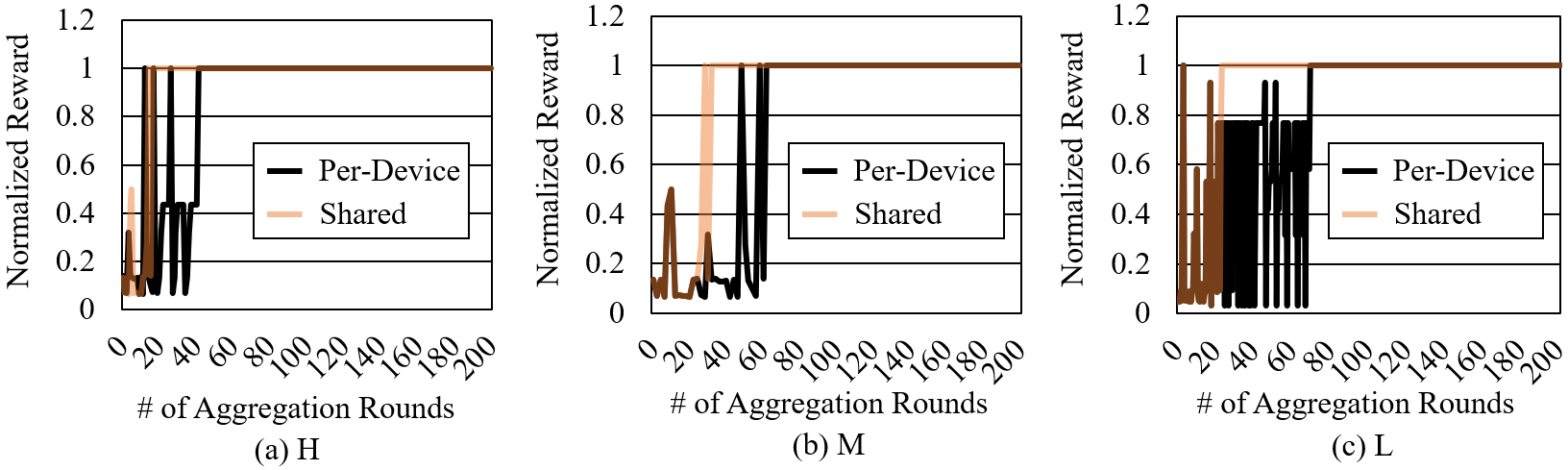}
    \vspace{-0.2cm}
    \caption{The reward is usually converged in 50-80 aggregations rounds. Sharing the Q-tables across the same category of devices expedites convergence.}
    \label{fig:overhead}
\end{figure}

Figure~\ref{fig:overhead} shows that, when training per-device Q-tables from scratch, the reward converges after about 50-80 aggregation rounds on average --- more than 200 rounds are usually required for FL convergence. Before convergence, AutoFL exhibits 28.3\% lower average energy efficiency than {\tt $O_{FL}$} due to the design space explorations. Nevertheless, it sill achieves 52.1\% energy saving against {\tt FedAvg-Random}. After the reward is converged, AutoFL accurately selects the participants and execution targets, as we observed in Section~\ref{sec:result2}.
As a result, AutoFL can achieve 5.2x energy efficiency improvement for the entire FL, on average.

The training overhead from the explorations can be alleviated by using the shared Q-tables. As shown in Figure~\ref{fig:overhead}, when the learned results are shared across the same category of devices, the training of RL converges more rapidly, reducing the average training overhead by 29.3\% --- the prediction accuracy of AutoFL is slightly degraded by 2.7\% though. This implies that, although each user experiences different degree of runtime variance and data heterogeneity, learned results from various devices complement one another.

The runtime cost of training per-device Q-tables is 531.5 $\mu$s, on average, excluding the time for FL execution. It corresponds to 0.8\% of the average time for aggregation rounds. The overhead consists of observing the per-device states (496.8 $\mu$s), selecting participants and execution targets based on the per-device Q-tables (10.5 $\mu$s), calculating the reward (2.1 $\mu$s), and updating the Q-tables (22.1 $\mu$s). The overhead from training computation can be further alleviated by leveraging idle cores in mobile SoCs --- the average thread level parallelism for mobile applications is around 2~\cite{CGao2015,YGKim2017_1} which is usually smaller than the number of available cores in mobile SoCs. Although AutoFL employs per-device Q-tables, the total memory requirement of AutoFL is feasible --- for our experiments with 200 devices, the total memory requirement of AutoFL is 80MB, 0.25\% of the typical 32GB DRAM capacity of commodity cloud servers.
During the inference phase, misprediction contributed negligible 5.6\% timing and 8.8\% energy efficiency overhead. AutoFL achieves an overall of 93.8\% prediction accuracy.
\section{Related Work}
\label{sec:related work}

\textbf{Energy Optimization for Mobile:} 
There are several prior works that proposed statistical models to capture uncertainties in the mobile environment for dynamic energy management~\cite{BGaudette2016,BGaudette2019,DShingari2018}.
For example, Gaudette et al. proposed to use arbitrary polynomial chaos expansions to consider the effect of various uncertainties on mobile user experience~\cite{BGaudette2019}.
Other computation offloading techniques also consider the performance variability aspect in the mobile environment for energy efficiency optimization~\cite{TAlfakih20,MAltamimi2015,ECuervo2010,LHuang2020,YGKim2017_2,YGKim2019,SPan2019,LQuan2018,BZhang2020,TZhang2019,YGKim2020}.
While the aforementioned techniques addressed similar runtime variance in the edge-cloud execution environment, prior works are sub-optimal for FL because of the highly distributed nature of FL use cases---not only that system and data heterogeneity can easily degrade the quality of FL, but runtime variance can also introduce uncertainties in FL's training time performance and execution efficiency.  

\textbf{Optimization for FL:} FL enables a large cluster of decentralized mobile devices at the edge to collaboratively train a shared ML model, while keeping the raw training samples on device~\cite{KBonawitz2019,RCGeyer2017,SItahara2020,JKonecny2016,ALi2020,WYBLim2020,HBMcMahan2017,ZTao2018,CWang2021,YZhan2020,JZhang2020}. To enable efficient FL deployment at the edge, FedAvg has been considered as the de facto FL algorithm~\cite{JKonecny2016,HBMcMahan2017}, which maximizes the computation-communication ratio by having less number of participant devices with higher per-device training iterations~\cite{HBMcMahan2017, VSmith2017, TLi2020}. On top of FedAvg, various works have been proposed to improve the accuracy of trained models~\cite{MDuan2021,ALi2020,TLi2020_2} or security robustness~\cite{RCGeyer2017,YLi2018,YLu2020}. While FedAvg has opened up the possibility for practical FL deployment, there are key optimization challenges.

The high degree of system heterogeneity and stochastic edge-cloud execution environment introduces the straggler problem in FL, where training time of each aggregation round is gated by the slowest device. To mitigate the straggler problem, previous works proposed to exclude stragglers from aggregation rounds~\cite{HBMcMahan2017,MMohri2019} or allow asynchronous update of gradients~\cite{YChen2019,MVDijk2020}. However, the aforementioned approaches often result in accuracy loss, because of insufficient gradient updates. 
On the other hand, Zhan et al. tried to exploit the stragglers for power saving by adjusting the CPU frequency only~\cite{YZhan2020}. However, their proposed technique can lead to significant increase in the overall training time. 

Varying characteristics of training samples per device introduce additional challenges to FL optimization. In particular, devices with non-IID data can significantly degrade model quality and convergence time~\cite{ZChai2019,XLi2020}. To mitigate the effect of data heterogeneity, previous approaches proposed to exclude updates from non-IID devices with asynchronous aggregation algorithms~\cite{YChen2019,YChen2019_2}, to warm up the global model with a subset of globally shared data~\cite{YZhao2018}, or to share data across a subset of devices~\cite{MDuan2021,ALi2020}. 
However, none of the aforementioned techniques explicitly take into account the stochastic runtime variance observed at the edge while handling data and system heterogeneity at the same time. 
Another FedAvg-based algorithm, called FedProx, was proposed~\cite{TLi2020}. FedProx handles system and data heterogeneity by allowing partial updates from stragglers and from participating devices with non-IID training data distribution.
However, since FedProx applies the same partial update rate to the randomly selected participants, it still does not deal with the heterogeneity, and stochastic runtime variance that can come from those randomly selected participants. In practice, AutoFL can be used with FedProx for improving the device selection approach.

Finally, there has been very little work on energy efficiency optimization for FL. Most prior work assume that FL is only activated when smartphones are plugged into wall-power~\cite{YChen2019_2,XQiu2020,ZXu2019,CWang2021,YZhan2020} limiting the practicality and adoption of FL.
To the best of our knowledge, AutoFL is the first work that demonstrates the potential of energy-efficient FL execution in the presence of realistic in-the-field effects: system and data heterogeneity with sources of performance uncertainties. By customizing a reinforcement learning-based approach, AutoFL can accurately identify an optimal cluster of participant devices and respective execution targets, adapting to heterogeneity and runtime variance.

\section{Conclusion}
\label{sec:conclusion}

Federated Learning has shown great promises in various applications with security-guarantee.
To enable energy efficient FL on energy-constrained mobile devices, we propose an adaptive, light-weight framework --- AutoFL.
The in-depth characterization of FL in edge-cloud systems demonstrates that an optimal cluster of participants and execution targets depend on various features: FL use cases, device and data heterogeneity, and runtime variance.
AutoFL continuously learns and identifies an optimal cluster of participant devices and their respective execution targets by taking into account the aforementioned features.
We design and construct representative FL use cases deployed in an emulated mobile cloud execution environment using off-the-shelf systems.
On average, AutoFL improves FL energy efficiency by 5.2x, compared to the baseline setting of random selection, while improving convergence time and accuracy at the same time.
We demonstrate that AutoFL is a viable solution and will pave the path forward by enabling future work on energy efficiency improvement for FL in realistic execution environment.  



\bibliographystyle{IEEEtranS}
\bibliography{refs}

\end{document}